\documentclass[11pt,letterpaper]{article}
\usepackage[letterpaper,margin=1in,top=0.95in,bottom=1in]{geometry}
\usepackage{fontspec}
\usepackage{amsmath,amssymb}
\usepackage{graphicx}
\usepackage{booktabs,tabularx,array}
\usepackage{xcolor}
\usepackage{microtype}
\usepackage{caption}
\usepackage{float}
\usepackage{listings}
\usepackage{enumitem}
\usepackage{xurl}
\usepackage{csquotes}
\usepackage{setspace}
\usepackage{fancyhdr}
\usepackage[backend=biber,style=numeric-comp,sorting=none,maxbibnames=12,doi=true,isbn=false,url=false,giveninits=true]{biblatex}
\usepackage[colorlinks=true,linkcolor=black,citecolor=black,urlcolor=blue]{hyperref}
\hypersetup{pdftitle={RIVER: Evidence-guided multi-agent retrieval for enzyme--reaction discovery},pdfauthor={Yifan Chang, Yunze Li, Feifei Jiang, Juntao Wu, Xin Chen, Han Hu, Yucheng Wu, Wei Tang, Mingqian Ma, Wanli Ouyang, Kaipeng Zhang, Lei Bai, Hongting Tang, Feiran Li, Yuchen Ren}}
\title{\vspace{-1.45em}\bfseries\LARGE RIVER: Evidence-guided multi-agent retrieval\\[0.1em] for enzyme--reaction discovery}
\author{%
  \begin{minipage}{0.95\textwidth}\centering
  \normalsize
  Yifan Chang\textsuperscript{1,2,3}, Yunze Li\textsuperscript{2},
  Feifei Jiang\textsuperscript{2,6}, Juntao Wu\textsuperscript{2},
  Xin Chen\textsuperscript{2}, Han Hu\textsuperscript{2,7},\\[0.35em]
  Yucheng Wu\textsuperscript{2,7}, Wei Tang\textsuperscript{2},
  Mingqian Ma\textsuperscript{2,8}, Wanli Ouyang\textsuperscript{2,9},\\[0.35em]
  Kaipeng Zhang\textsuperscript{1}, Lei Bai\textsuperscript{2},
  Hongting Tang\textsuperscript{5}, Feiran Li\textsuperscript{4},
  Yuchen Ren\textsuperscript{2,*}
  \\[0.9em]
  \small
  \textsuperscript{1}Shanghai Innovation Institute
  \quad \textsuperscript{2}Shanghai AI Lab\\[0.25em]
  \textsuperscript{3}University of Science and Technology of China
  \quad \textsuperscript{4}Tsinghua University\\[0.25em]
  \textsuperscript{5}Sun Yat-sen University
  \quad \textsuperscript{6}Shanghai Jiao Tong University\\[0.25em]
  \textsuperscript{7}Fudan University
  \quad \textsuperscript{8}Carnegie Mellon University\\[0.25em]
  \textsuperscript{9}The Chinese University of Hong Kong\\[0.45em]
  \textsuperscript{*}Corresponding author: Yuchen Ren (\href{mailto:renyuchen@pjlab.org.cn}{renyuchen@pjlab.org.cn})
  \end{minipage}
}
\date{}
\begin{document}
\maketitle
\vspace{-1.5em}
% Preserve all 50 source references in their original numeric order.
% Some are bibliographic background materials not explicitly cited in this source draft.
\nocite{*}
\begin{abstract}
Enzyme function annotation and the identification of candidate enzymes for a given reaction are essential steps in connecting genomic information to biochemical reaction networks. However, existing methods typically rely on sequence or structural similarity or on training with known enzyme--reaction pairs, making it difficult to integrate existing biochemical evidence across large candidate spaces; they also struggle with retrieval tasks involving enzyme families whose members are highly similar. Here, we introduce RIVER (Reaction--enzyme Inference via Evidence-guided Retrieval), a multi-agent framework for bidirectional enzyme--reaction retrieval that identifies potentially catalyzed reactions for a given enzyme and ranks candidate enzymes for a given reaction. RIVER coordinates large language models, specialized biochemical computational tools, and external knowledge resources to assemble evidence at multiple levels, including reaction chemistry, protein sequences and domains, catalytic mechanisms, and database annotations. Dedicated validation modules further examine evidence gaps and potential conflicts and rerank candidates using traceable evidence records. On public benchmarks, RIVER achieves the highest success rates in both enzyme retrieval and enzyme annotation. In family-specific enzyme retrieval, RIVER outperforms the evaluated baselines on the selected datasets and metrics. In \emph{Escherichia coli} enzyme retrieval tasks constructed from recent literature, RIVER achieves the best retrieval results with both closed-source and open-source model configurations, substantially outperforming closed-source general-purpose and biology-specialized multi-agent systems, including Claude Code, Codex, and Phylo Max. These results suggest that evidence-guided multi-agent retrieval can prioritize biochemically supported enzyme--reaction relationships, providing a basis for subsequent experimental validation.
\end{abstract}

\section{Introduction}\label{introduction}

Enzyme--reaction relationships provide mechanistic links between genomic information and biochemical transformations. They underpin protein function annotation, genome-scale metabolic reconstruction, pathway analysis, and biocatalyst discovery for chemical and industrial applications ~\cite{ref1,ref2}. However, the increasing abundance of protein sequences and metabolomics data has not eliminated the functional annotation bottleneck: many proteins remain insufficiently characterized, and standard annotations may not adequately capture the breadth of catalytic promiscuity or substrate-specific activities ~\cite{ref3,ref4,ref5,ref6,ref7}. Experimental metabolomics screens have begun to reveal previously hidden enzyme activities at scale, while also highlighting the challenge of prioritizing informative candidates for mechanistic follow-up studies ~\cite{ref8,ref9}.

Computational approaches to enzyme function encompass sequence homology and annotation transfer, sequence-based enzyme classification, structure-informed prediction, and reaction-conditioned retrieval. Similarity searches and pathway-oriented enzyme selection remain valuable for locating relevant catalysts ~\cite{ref10,ref11}; DeepEC and CLEAN infer Enzyme Commission (EC) functions from sequences ~\cite{ref12,ref13}, whereas graph-based and geometric approaches use structural information to predict function or catalytic classes ~\cite{ref14,ref15}. Methods that explicitly represent substrates or reactions, including ESP, CLIPZyme, and EnzymeMap, address chemical compatibility more directly ~\cite{ref16,ref17,ref18}. Recent work such as EnzymeCAGE and EnzGFM has advanced enzyme retrieval and enzyme-oriented foundation model research, whereas RXNRECer and ReactZyme emphasize fine-grained enzyme--reaction annotation and reaction-level evaluation ~\cite{ref19,ref20,ref21,ref22}. These methods address overlapping but distinct tasks, and existing approaches should not be uniformly reduced to similarity matching. Nevertheless, a predicted label or ranking score does not necessarily distinguish directly documented catalytic activity from mechanistic analogy or unresolved substrate specificity, particularly when different methods use different candidate pools or evaluation protocols. This distinction matters when computational predictions are used to prioritize candidates for biochemical research.

Evidence relevant to enzyme--reaction hypotheses is distributed across resources with different scopes. UniProtKB records protein annotations, Rhea curates chemically well-defined biochemical transformations, and BRENDA compiles enzyme properties ~\cite{ref23,ref24,ref25}; protein language models and structural resources provide complementary information, especially for poorly annotated proteins ~\cite{ref26,ref27,ref28,ref29,ref30}. However, the availability and specificity of these records vary. The absence of annotation does not imply the absence of activity, just as a chemically plausible mechanism does not establish that a particular protein actually catalyzes the queried reaction. Retrieval-augmented language models and tool-using agents offer ways to integrate dispersed information ~\cite{ref31,ref32}; chemistry-oriented agents and scientific multi-agent systems have demonstrated tool use, hypothesis generation, and critical review in other settings ~\cite{ref33,ref34,ref35,ref36,ref37}. In enzyme research, EnzGFM-Agent provides a relevant precedent for agent-based approaches ~\cite{ref20}. Thus, an important methodological opportunity is not simply to combine multiple agents, but to organize evidence acquisition, biochemical interpretation, review of missing or potentially conflicting support, and candidate reranking into a unified and inspectable retrieval workflow. Such a workflow should make the reasons for candidate prioritization traceable, while avoiding the conflation of supporting evidence with experimental confirmation of catalytic activity.

\begin{figure}[p]
\centering
\includegraphics[width=\linewidth,height=0.60\textheight,keepaspectratio]{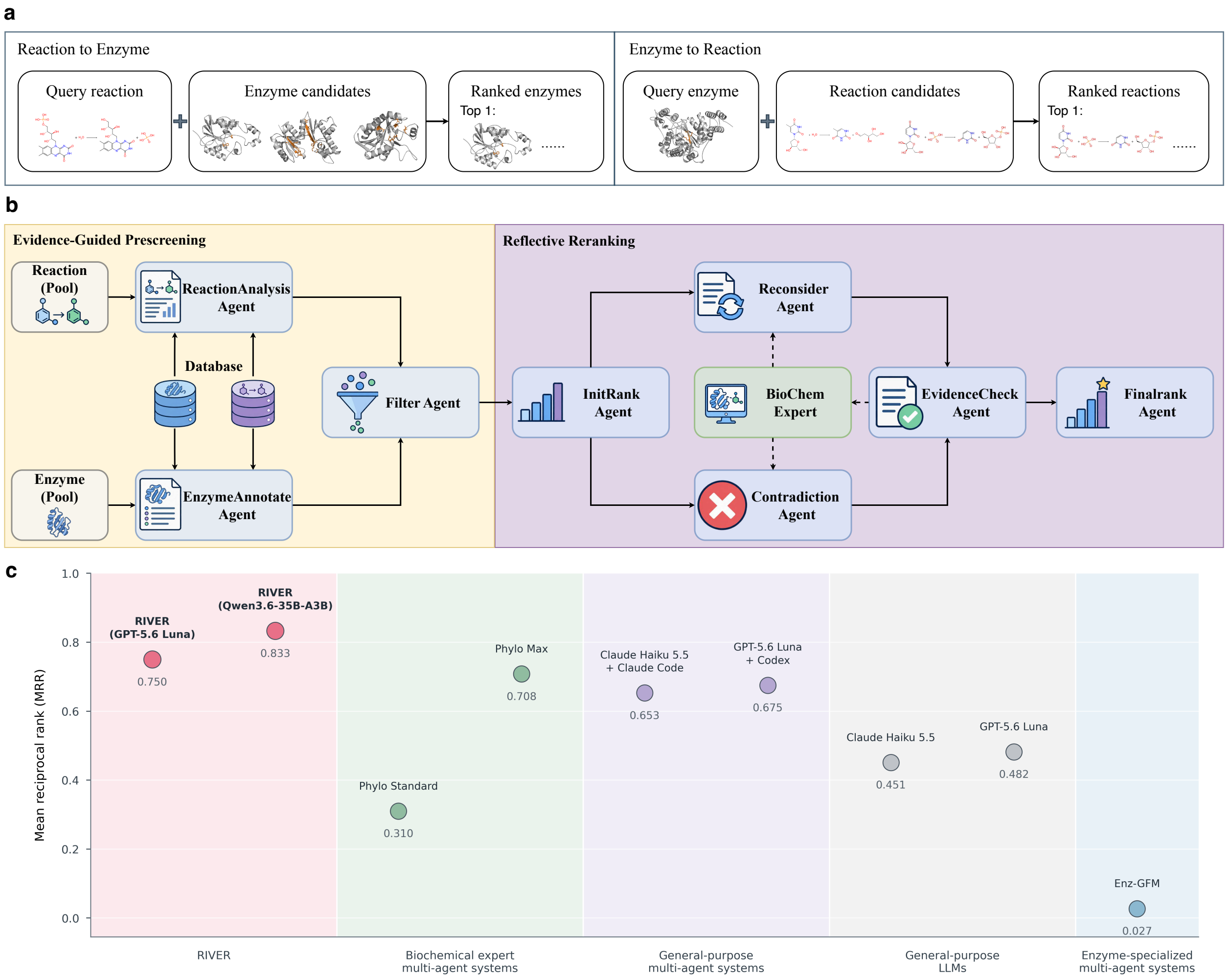}
\caption{\textbf{RIVER: an evidence-guided multi-agent framework for bidirectional enzyme--reaction retrieval.} \textbf{a, Complementary retrieval settings.} Given a query reaction, RIVER ranks candidate enzymes; given a query enzyme, it ranks candidate reactions. \textbf{b, Two-stage workflow.} During evidence-guided prescreening, the ReactionAnalysis and EnzymeAnnotate agents compile reaction-level and protein-level information with database support, while the Filter Agent narrows the candidate pool. During reflective reranking, the InitRank Agent produces an initial order, the Reconsider and Contradiction agents reassess supporting and conflicting evidence, and BioChem Expert provides computational evidence. The EvidenceCheck Agent reviews the resulting evidence, after which the FinalRank Agent generates the final ranking. Dashed connections indicate interactions involving computational expertise. \textbf{c, Comparison with multi-agent systems.} We constructed \emph{E. coli} enzyme retrieval tasks from recent literature and compared RIVER with different categories of state-of-the-art multi-agent systems, obtaining a clear performance advantage.}
\label{fig:framework}
\end{figure}

Here, we present RIVER (Reaction--enzyme Inference via Evidence-guided Retrieval), a training-free multi-agent framework that integrates bidirectional enzyme--reaction retrieval through a unified evidence-guided approach. The two directions address complementary scientific needs: enzyme-to-reaction retrieval supports protein function annotation and metabolic network reconstruction, whereas reaction-to-enzyme retrieval supports the identification of candidate biocatalysts for specific chemical transformations (Fig.~\ref{fig:framework}a). RIVER coordinates large language models (LLMs), external knowledge resources, and biochemical computational tools to assemble heterogeneous evidence spanning reaction chemistry, protein sequences and domains, catalytic mechanisms, and database annotations. Dedicated review modules examine evidence gaps and potential conflicts and use the resulting assessments to inform candidate reranking, ultimately producing ranked candidates with traceable, candidate-specific rationales (Fig.~\ref{fig:framework}b). In \emph{E. coli} enzyme retrieval tasks constructed using data and findings from recent literature, our multi-agent system accurately retrieves enzymes corresponding to biochemical reactions, outperforming general-purpose and biochemistry-specialized multi-agent systems, including Claude Code, Codex, and Phylo Max, while demonstrating robustness to the choice of foundation model (Fig.~\ref{fig:framework}c). On public benchmarks, we substantially outperform existing baselines in both enzyme retrieval and enzyme annotation; our multi-agent system also demonstrates a clear advantage in the challenging setting of family-specific enzyme retrieval.

\section{Results}\label{results}

\subsection{\texorpdfstring{RIVER in experimentally validated \emph{E. coli} enzyme discovery reported in the literature}{RIVER in experimentally validated E. coli enzyme discovery reported in the literature}}\label{river-in-experimentally-validated-e.-coli-enzyme-discovery-reported-in-the-literature}

\begin{figure}[p]
\centering
\includegraphics[width=\linewidth,height=0.60\textheight,keepaspectratio]{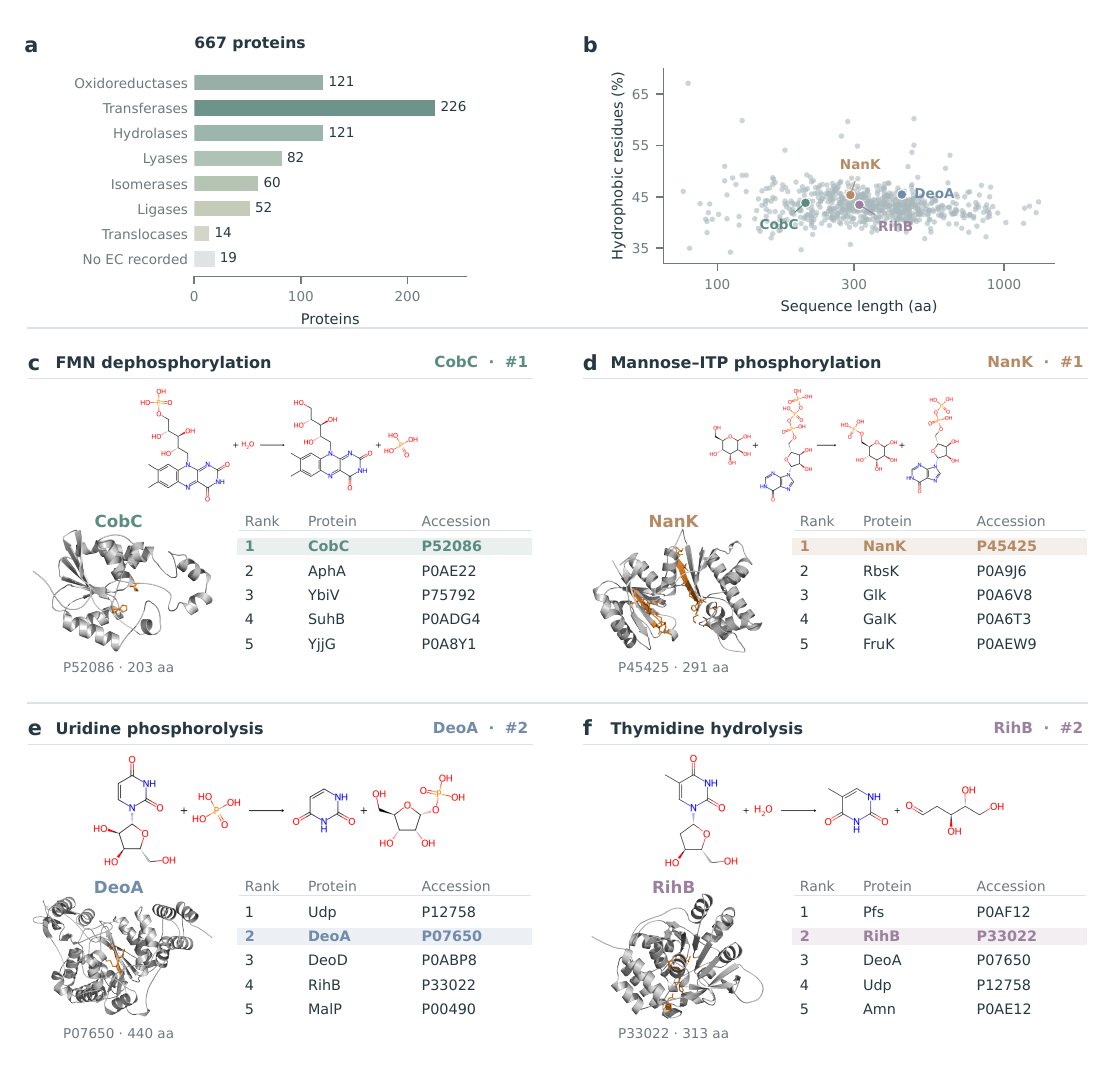}
\caption{\textbf{RIVER prioritizes experimentally validated enzyme--reaction associations among 667 \emph{E. coli} proteins.} \textbf{a,} Counts of the 667 candidate proteins by primary EC class; classes are not mutually exclusive, and 19 proteins lack recorded EC annotations. \textbf{b,} Candidate protein sequence lengths and proportions of hydrophobic residues (A, V, I, L, M, F, W, Y); colored points indicate the four literature reference enzymes. \textbf{c--f,} Reaction structures, reference protein structures, and top-five candidates for FMN dephosphorylation (CobC, P52086), mannose--ITP phosphorylation (NanK, P45425), uridine phosphorolysis (DeoA, P07650), and thymidine hydrolysis (RihB, P33022). Colored rows indicate the literature reference enzymes; other candidates are not presumed to lack catalytic activity. Protein structures are shown as post hoc visualization context and were not inputs to structure-pocket tools during ranking. Panels c--f correspond to the GPT-5.6 Luna configuration.}
\label{fig:ecoli}
\end{figure}

Discovering enzyme functions requires more than inferring broad catalytic classes from protein sequences; it also requires identifying enzymes capable of performing specific biochemical transformations among numerous candidates with similar functions. To first evaluate the practical value of RIVER in this research setting, we used the \emph{Escherichia coli} in vitro metabolomics study reported by Gruber \emph{et al.}~\cite{ref9} in \emph{Communications Biology} (2026) as a retrospective case study. Through experimental screening of 667 successfully purified metabolic enzymes from the ASKA collection~\cite{ref44}, that study identified multiple previously undercharacterized catalytic activities and further validated selected reactions experimentally. We selected four experimentally supported enzyme--reaction associations: CobC-catalyzed dephosphorylation of FMN to riboflavin, NanK-catalyzed phosphorylation of mannose using ITP, DeoA-catalyzed uridine phosphorolysis, and RihB-catalyzed thymidine hydrolysis. Rather than comparing only a small number of homologous proteins, we ranked the same 667 candidate proteins for each reaction, yielding a total of 2,668 reaction--enzyme candidate associations (Fig.~\ref{fig:ecoli}a--f). This setting directly tests whether the system can prioritize the experimentally confirmed extensions of substrate or reaction scope reported in the literature using existing biochemical knowledge and incomplete protein annotations.

The 667 candidate proteins span all seven primary EC classes, including 226 proteins in the transferase class and 121 each in the oxidoreductase and hydrolase classes; another 19 proteins have no recorded EC number. Counts across EC classes are not mutually exclusive: among the 648 proteins with EC annotations, 25 belong to more than one primary class (Fig.~\ref{fig:ecoli}a). Candidate sequence lengths range from 76 to 1,320 amino acid residues, with a median of 336, and the proportion of hydrophobic residues also varies (Fig.~\ref{fig:ecoli}b). Thus, the candidate pool includes both proteins with related functions identifiable from reaction classes and candidates requiring finer distinctions based on substrate structures, cosubstrate selection, and chemical bond changes.

We compared RIVER against direct LLM calls, combinations involving general-purpose coding agents, and existing specialized retrieval methods (Table~\ref{tab:ecoli}). With \textbf{Qwen3.6-35B-A3B}~\cite{qwen36_2026} as the foundation model, RIVER ranked CobC, NanK, and RihB first and DeoA third, achieving a mean reciprocal rank (MRR) of \textbf{0.833} across the four cases. With \textbf{GPT-5.6 Luna}~\cite{openai_gpt56_2026}, the final ranks of the four target enzymes were \textbf{1, 1, 2, and 2}, respectively, all within the top two candidates, with an MRR of \textbf{0.750} (Fig.~\ref{fig:ecoli}c--f). The two configurations achieved 3/4 and 2/4 Top-1 hits, respectively, and both achieved 4/4 Top-3 hits. Although their rankings differed for individual substrates, both effectively narrowed the range of candidates meriting priority investigation.

\begin{table}[H]
\centering
\caption{\textbf{Retrieval performance of RIVER, general-purpose coding agents, language models, and existing enzyme retrieval methods on four experimentally validated reactions.} Each column corresponds to a query reaction and its literature reference enzyme. Values represent the rank of the reference enzyme in the candidate pool; lower is better. MRR is the mean reciprocal rank across the four queries; higher is better. For configurations with variation across three repeated runs, mean ± standard deviation is reported; the absence of a reported standard deviation indicates identical results across all three runs.}
\label{tab:ecoli}
\vspace{4pt}
\begingroup\footnotesize\setlength{\tabcolsep}{3.5pt}\renewcommand{\arraystretch}{1.35}
\begin{tabularx}{\linewidth}{@{}>{\raggedright\arraybackslash}X *{4}{>{\centering\arraybackslash}p{1.52cm}}>{\centering\arraybackslash}p{2.10cm}>{\centering\arraybackslash}p{1.72cm}>{\centering\arraybackslash}p{1.72cm}>{\centering\arraybackslash}p{1.72cm}>{\centering\arraybackslash}p{1.72cm}>{\centering\arraybackslash}p{1.72cm}>{\centering\arraybackslash}p{1.72cm}>{\centering\arraybackslash}p{1.72cm}>{\centering\arraybackslash}p{1.72cm}>{\centering\arraybackslash}p{1.72cm}>{\centering\arraybackslash}p{1.72cm}@{}}
\toprule
\textbf{System} & \textbf{FMN / CobC} & \textbf{Mannose--ITP / NanK} & \textbf{Uridine / DeoA} & \textbf{Thymidine / RihB} & \textbf{MRR $\uparrow$} \\
\midrule
\textbf{RIVER (Qwen3.6-35B-A3B)} & \textbf{1} & \textbf{1} & 3 & \textbf{1} & \textbf{0.833} \\
\textbf{RIVER (GPT-5.6 Luna)} & \textbf{1} & \textbf{1} & \textbf{2} & 2 & \textbf{0.750} \\
\midrule
Phylo Max & 1 & 2 & 3 & 1 & 0.708 \\
GPT-5.6 Luna + Codex & 1 & 5 & 2 & 1 & 0.675 \\
Claude Haiku 5.5 + Claude Code & 1.7  $\pm$  1.2 & 2.3  $\pm$  0.6 & 2.7  $\pm$  0.6 & 1.0  $\pm$  0.0 & 0.653  $\pm$  0.105 \\
GPT-5.6 Luna & 3.3  $\pm$  1.5 & 2.7  $\pm$  1.2 & 2.0  $\pm$  0.0 & 1.7  $\pm$  0.6 & 0.482  $\pm$  0.112 \\
Claude Haiku 5.5 & 5.3  $\pm$  1.2 & 2.3  $\pm$  1.2 & 2.3  $\pm$  0.6 & 2.0  $\pm$  1.0 & 0.451  $\pm$  0.103 \\
Phylo Standard & 14 & 3 & 2 & 3 & 0.310 \\
Enz-GFM & 22 & 105 & 339 & 20 & 0.027 \\
\bottomrule
\end{tabularx}
\endgroup
\end{table}

In this comparison, \textbf{both RIVER configurations achieved higher MRRs than all control settings listed in the table}. Their advantage was evident not only over stand-alone language models but also over general-purpose agents capable of planning, code execution, and external tool calls. Claude Haiku 5.5~\cite{anthropic_haiku55_2026} combined with Claude Code~\cite{anthropic_claudecode} achieved an MRR of \textbf{0.653 ± 0.105}, GPT-5.6 Luna combined with Codex~\cite{openai_codex_2025} achieved \textbf{0.675}, and Phylo Max~\cite{phylo_platform_2026} achieved \textbf{0.708}; all were below both RIVER (GPT-5.6 Luna), at \textbf{0.750}, and the Qwen-based version, at \textbf{0.833}. Under the comparison using the same GPT-5.6 Luna foundation model, RIVER exceeded Luna + Codex by \textbf{0.075} MRR and the mean from repeated stand-alone Luna API experiments (\textbf{0.482 ± 0.112}) by \textbf{0.268}. Thus, for this retrieval task defined by specific catalytic reactions, the programming and tool-use capabilities of general-purpose agents do not necessarily translate into optimal enzyme--reaction matching; a dedicated workflow for biochemical evidence analysis, review, and ranking provides additional value.

Differences from existing enzyme retrieval reference methods were similarly pronounced. Phylo Standard~\cite{phylo_platform_2026} achieved an MRR of \textbf{0.310} across the four tasks; the literature-based comparison result for Enz-GFM~\cite{ref20} was \textbf{0.027}, with NanK and DeoA ranked 105th and 339th, respectively. For reactions that extend beyond conventional functional annotations to varying degrees, the retrieval challenge is not merely to identify a broad enzyme class but to assess whether a candidate may accept a noncanonical substrate or cosubstrate and perform the required chemical transformation at the target site.

\subsubsection{Different substrates and cosubstrates pose specificity challenges for candidate enzymes}\label{different-substrates-and-cosubstrates-pose-specificity-challenges-for-candidate-enzymes}

The four experimental cases offer complementary perspectives on the basis of RIVER's rankings (Fig.~\ref{fig:ecoli}c--f). For \textbf{CobC}, known activities related to adenosylcobalamin phosphate and α-ribazole phosphate, together with histidine phosphatase family information, provide an analogical basis for its potential dephosphorylation of FMN, another vitamin-derived phosphate ester. The key inference here is an extrapolation from shared phosphate-ester hydrolysis chemistry to a molecule with a substantially different substrate scaffold, rather than a simple match to an identical substrate name. For \textbf{NanK}, the canonical function is the phosphorylation of N-acetylmannosamine. The system combines information about the C6 reaction position of hexoses with related sugar kinase evidence to support the possibility that NanK accepts mannose, although the retrieved records do not directly establish that ITP can replace the usual phosphate donor. The follow-up experimental validation in the original metabolomics study and the annotation evidence available during RIVER retrieval represent distinct levels of information and should be distinguished.

For \textbf{DeoA}, its pyrimidine nucleoside phosphorolysis chemistry is compatible with the phosphate-dependent cleavage of the glycosidic bond and formation of ribose-1-phosphate in the uridine reaction. RIVER (GPT-5.6 Luna) ranked DeoA second, while the known uridine phosphorylase Udp ranked first. This outcome shows that the system retained the substrate-scope extension reported in the literature without excluding other enzymes already capable of catalyzing the reaction. For \textbf{RihB}, its pyrimidine nucleoside hydrolase function provides important evidence at the reaction-type level, but a substrate-specificity difference remains to be explained between previously documented ribonucleoside activity and the 2$^{\prime}$-deoxyribose structure of the target substrate, thymidine. Together, these cases show that the practical value of the system lies both in placing the experimental reference enzymes within a high-priority range for validation and in preserving specific evidence gaps concerning substrates, cosubstrates, and structural selectivity.

\subsection{Competitive performance of RIVER on public benchmarks}\label{competitive-performance-of-river-on-public-benchmarks}

\subsubsection{Benchmark datasets}\label{benchmark-datasets}

To systematically evaluate the generalization of RIVER in bidirectional enzyme--reaction retrieval and across different enzyme families, we used five public benchmark datasets provided by EnzymeCAGE~\cite{ref19}: two internal test sets constructed from the Rhea database (Enzyme-405 and Orphan-335) and three external test sets targeting specific enzyme families (P450, Terpene, and Phosphatase).

\textbf{Enzyme-405} primarily evaluates functional prediction for unseen enzymes. It contains 405 enzymes and 295 biochemical reactions, forming 961 positive enzyme--reaction pairs and 14,960 negative pairs. \textbf{Orphan-335} contains 335 reactions absent from the training data but subsequently assigned enzymatic annotations in later database versions, and evaluates the ability to retrieve known candidate enzymes for a target reaction. Both datasets were constructed from annotations added to Rhea between July 2023 and February 2025, with reaction-similarity filtering applied to reduce overlap with the training data.

The three external test sets further examine retrieval performance in enzyme families with different biochemical characteristics. The \textbf{P450} dataset originates from the manually curated P450Rdb and covers oxidative reactions catalyzed by cytochrome P450 enzymes from multiple species, emphasizing differences in substrate structures, oxidation sites, and regioselectivity. The \textbf{Terpene} dataset contains terpene synthases and their catalyzed reactions, involving cyclization and scaffold rearrangement of different isoprenoid precursors, and tests the ability to distinguish similar substrates and different product scaffolds. The \textbf{Phosphatase} dataset comes from a high-throughput enzyme activity screening study; the original data include 218 phosphatases and 165 reactions, totaling 35,970 enzyme--reaction combinations, and focus on hydrolytic specificity toward different substrates within the same family.

Together, these five datasets cover functional prediction for unseen enzymes, candidate enzyme retrieval for unseen reactions, and the identification of substrate and reaction specificity within different catalytic families. They provide complementary test settings for assessing RIVER's bidirectional retrieval performance and biochemical generalization.

\begin{figure}[p]
\centering
\includegraphics[width=\linewidth,height=0.60\textheight,keepaspectratio]{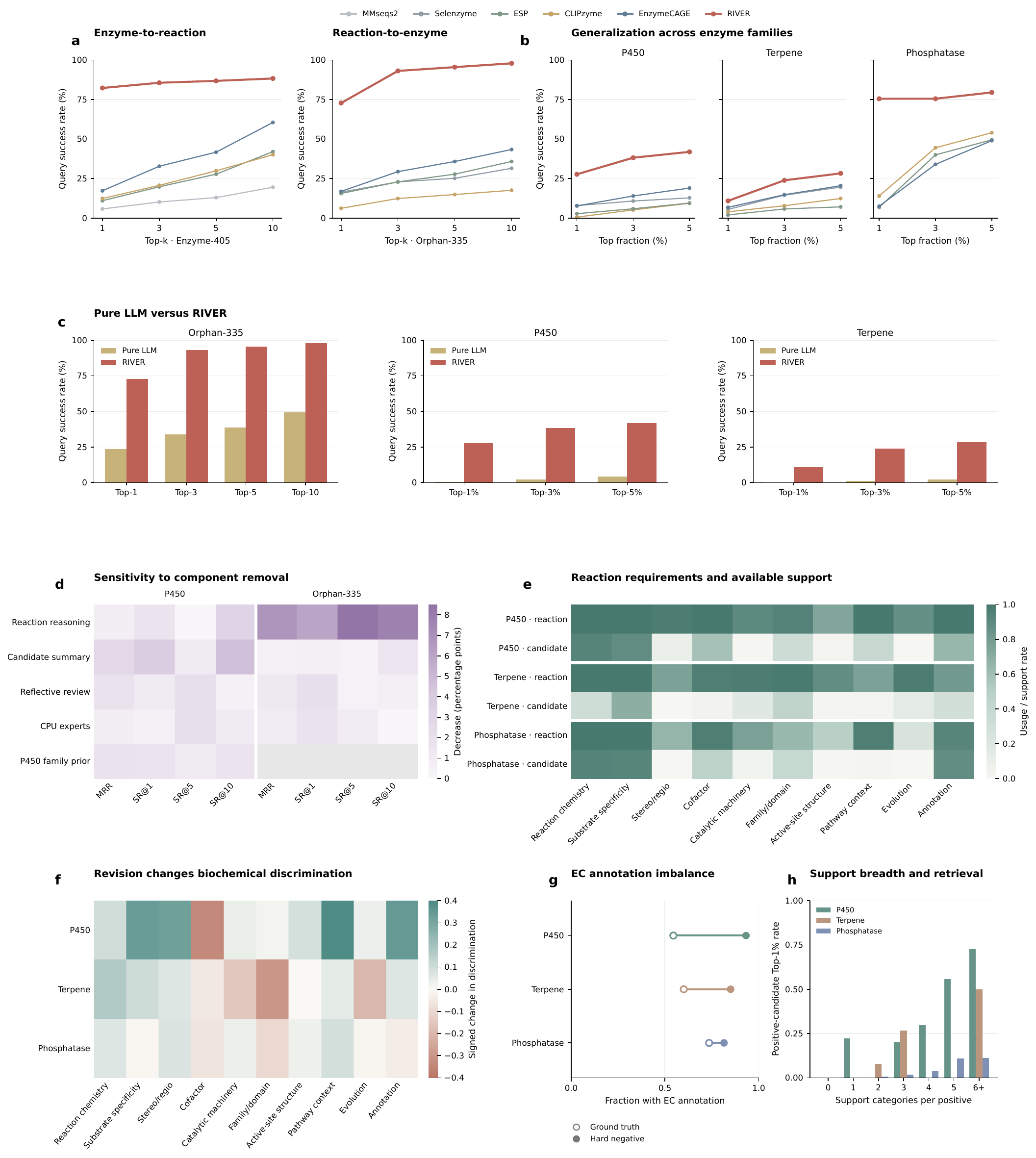}
\caption{\textbf{RIVER's bidirectional enzyme--reaction retrieval performance, module ablations, and biochemical evidence analysis.} \textbf{a,} E2R retrieval on Enzyme-405 and R2E retrieval on Orphan-335; the curves show the fraction of queries with at least one reference-positive candidate within the Top-\emph{k} ranks. \textbf{b,} Query success rates for reference positives entering the top 1\%, 3\%, and 5\% of candidate pools in the P450, Terpene, and Phosphatase datasets. \textbf{c,} Comparison between the multi-agent system using a locally deployed LLM (GLM-5.2 as its base model) and ranking using GLM-5.2 alone without online search. \textbf{d,} Decreases in MRR and Top-1, Top-5, and Top-10 success rates relative to the full system following the removal of reaction reasoning, candidate summarization, reflection review, CPU experts, and family priors. Gray indicates that a module is not applicable; MRR differences are plotted on a percentage scale. \textbf{e,} Reaction-side requirements (upper row in each group) and coverage of supporting evidence among the top ten candidates (lower row) for ten biochemical attributes in each family. \textbf{f,} Changes in net discrimination between reference positives and hard negatives across biochemical evidence categories before and after reassessment; positive values indicate increased discrimination and negative values indicate decreased discrimination. \textbf{g,} EC annotation coverage for reference positives (open circles) and hard negatives (filled circles). \textbf{h,} Proportions of reference-positive candidates entering the top 1\% of candidate pools, grouped by the number of supporting evidence categories.}
\label{fig:benchmarks}
\end{figure}

To test whether the advantages observed in the case studies extend to standardized retrieval tasks, we next examined both enzyme-to-reaction (E2R) and reaction-to-enzyme (R2E) retrieval and compared RIVER against baselines including MMseqs2, Selenzyme, ESP, CLIPzyme, and EnzymeCAGE (Fig.~\ref{fig:benchmarks}a--b). On Enzyme-405, RIVER ranked candidates for 402 evaluable queries out of 405, achieving \textbf{Top-1, Top-3, Top-5, and Top-10} query success rates of \textbf{82.3\%, 85.6\%, 86.8\%, and 88.3\%}, respectively. Its Top-1 success rate exceeded the \textbf{17.3\%} shown for EnzymeCAGE in the figure (Fig.~\ref{fig:benchmarks}a). On this task, MRR was \textbf{0.846}, MAP was \textbf{0.831}, nDCG@10 was \textbf{0.841}, and the enrichment factor within the top 1\% of candidates was \textbf{84.1}. These results indicate that RIVER can not only rank candidate proteins for a given reaction but also prioritize the recovery of corresponding biochemical reactions from enzyme information.

On Orphan-335 reaction-to-enzyme retrieval, RIVER achieved Top-1, Top-3, Top-5, and Top-10 query success rates of \textbf{72.8\%, 93.1\%, 95.5\%, and 97.9\%}, respectively (Fig.~\ref{fig:benchmarks}a). The corresponding EnzymeCAGE Top-1 value was \textbf{16.9\%}. As with Enzyme-405, the high Top-1 success rate indicates that the system can place known associations first among numerous reference candidates, while the further improvement at Top-10 shows that reference positives can still be recovered within relatively high ranks for queries without a first-place hit.

We further evaluated success rates within the top 1\%, 3\%, and 5\% of candidate pools on three family-specific benchmarks: P450, terpene synthases, and phosphatases (Fig.~\ref{fig:benchmarks}b). RIVER achieved corresponding success rates of \textbf{27.7\%, 38.2\%, and 41.9\%} on P450; \textbf{10.9\%, 23.9\%, and 28.3\%} on Terpene; and \textbf{75.5\%, 75.5\%, and 79.5\%} on Phosphatase. At the stringent top-1\% threshold, these results represent gains of \textbf{19.8, 4.1, and 61.5 percentage points}, respectively, over the strongest comparator shown for each dataset. The considerable differences in absolute success rates and relative gains among the three families suggest that substrate classes, family annotation coverage, and candidate distributions affect the difficulty of reaction-conditioned retrieval.

\subsection{Evidence integration mechanisms in RIVER}\label{evidence-integration-mechanisms-in-river}

RIVER's performance depends not only on the final rankings, but also on whether prescreening retains the correct candidates and whether subsequent reasoning can extract discriminative reaction information. On P450, prescreening retained \textbf{51.1\%} of all candidate pairs while preserving \textbf{97.2\%} of labeled positive pairs; \textbf{96.9\%} of queries retained at least one positive. On Phosphatase, \textbf{72.3\%} of candidate pairs and \textbf{79.6\%} of positive pairs were retained, and every query retained at least one labeled positive. These results show that a lower candidate retention rate can coexist with high query-level positive coverage.

Effective online information retrieval is particularly important to the overall multi-agent system. Figure 3c compares the performance of a multi-agent system composed solely of LLMs with that of our multi-agent system. To exclude possible online-access capabilities of closed-source models, we used the locally deployed GLM-5.2 as the foundation model for both the LLM and our multi-agent system in this experiment. The results show that retrieval performance decreases markedly without online data and knowledge as prerequisites. This demonstrates that the biochemical knowledge and reasoning capabilities inherent in an open-source LLM alone are insufficient for this task, and that knowledge augmentation using existing database resources is needed.

To determine the relative contributions of different components, we separately removed reaction reasoning, candidate summarization, reflection review, and CPU biochemical computational experts, and examined changes in the final metrics relative to the full system (Fig.~\ref{fig:benchmarks}d). On Orphan-335, removing \textbf{reaction reasoning} reduced MRR by \textbf{0.068}, while Top-1, Top-5, and Top-10 success rates decreased by \textbf{6.0, 8.4, and 7.8 percentage points}, respectively. For orphan reactions lacking known direct catalytic representatives, functional names or sequence similarities alone often cannot reliably connect candidates to the target chemical transformation. Active interpretation of reactants, products, and transformation types is therefore an important source of information for prioritizing candidates.

By contrast, the largest MRR loss on P450 occurred when \textbf{candidate summarization} was removed: MRR decreased by \textbf{0.0268}, while Top-1 and Top-10 success rates declined by \textbf{3.7} and \textbf{4.7 percentage points}, respectively (Fig.~\ref{fig:benchmarks}d). P450 candidates share much of their enzyme-family background, and generic oxidase labels cannot readily distinguish particular substrate scaffolds and oxidation sites. Integrating candidate annotations, substrate ranges, reaction positions, and existing functional evidence is therefore particularly important. Reflection review and CPU experts also alter some rankings.

\subsection{Relationships among biochemical evidence gaps, selectivity, and candidate ranking}\label{relationships-among-biochemical-evidence-gaps-selectivity-and-candidate-ranking}

We further investigated how different types of biochemical evidence contribute to candidate ranking by comparing the constraints proposed on the reaction side with supporting evidence among the top ten candidates (Fig.~\ref{fig:benchmarks}e). Coverage differed substantially across the ten categories of attributes. For example, approximately 97\% of P450 reaction analyses involved stereo- or regioselectivity, whereas only approximately 8\% of the top ten candidates had corresponding supporting information. This indicates that the attributes relevant to the reactions and the supporting evidence commonly available for candidates have different distributions. In conjunction with the reassessment analysis, evidence concerning stereo- or regioselectivity in P450 showed greater discrimination between positives and hard negatives after reassessment (Fig.~\ref{fig:benchmarks}f), indicating that even evidence with relatively limited coverage may have discriminative value. Meanwhile, hard negatives had higher EC annotation coverage than true-positive enzymes (Fig.~\ref{fig:benchmarks}g), showing that relatively extensive functional annotations may also occur for mismatched candidates. These findings support assessing both evidence coverage and its ability to distinguish candidates with respect to the target reaction, with particular attention to information on substrates, reaction sites, and stereoselectivity that can differentiate similar candidates.

Further analysis of the net discriminatory ability of different attributes between reference positives and hard negatives before and after reassessment revealed nonuniform changes in attribute weighting (Fig.~\ref{fig:benchmarks}f). In P450, net discrimination associated with substrate specificity increased by approximately \textbf{0.335}, whereas that associated with cofactor evidence decreased by approximately \textbf{0.335}. This does not mean that cofactor information is unimportant for biochemical reactions. Rather, within the candidate set and comparison scheme considered, it was less capable of distinguishing known positives from hard negatives than the substrate-specificity information emphasized after revision. Accordingly, a more valuable role of reflection may be to reconsider whether evidence genuinely distinguishes competing candidates, rather than merely accumulating additional seemingly relevant facts.

A comparison of EC annotation coverage further shows that more extensive annotation does not necessarily indicate a better match to the target reaction (Fig.~\ref{fig:benchmarks}g). In all three families, hard negatives had higher EC annotation coverage than reference positives; for P450, the corresponding proportions were \textbf{93.2\%} and \textbf{54.5\%}, respectively. When ranking candidates, the system must go beyond asking whether a protein has an EC annotation to asking whether that annotation supports the target substrate, chemical bond changes, and reaction site. Correspondingly, the relationship between evidence breadth and retrieval success is family-dependent (Fig.~\ref{fig:benchmarks}h). Among P450 reference positives with five categories of supporting evidence, \textbf{55.7\%} entered the top 1\% of the candidate pool; among those with six or more evidence categories, the proportion increased to \textbf{72.7\%}. However, the proportions across evidence-count groups were not monotonically increasing for terpene synthases, and between-group differences were smaller for phosphatases. This suggests that the \textbf{specificity and discriminative power} of evidence deserve greater attention than its simple count.

\subsection{Six cases across reaction types reveal traceable evidence-guided reranking}\label{six-cases-across-reaction-types-reveal-traceable-evidence-guided-reranking}

\begin{figure}[p]
\centering
\includegraphics[width=\linewidth,height=0.60\textheight,keepaspectratio]{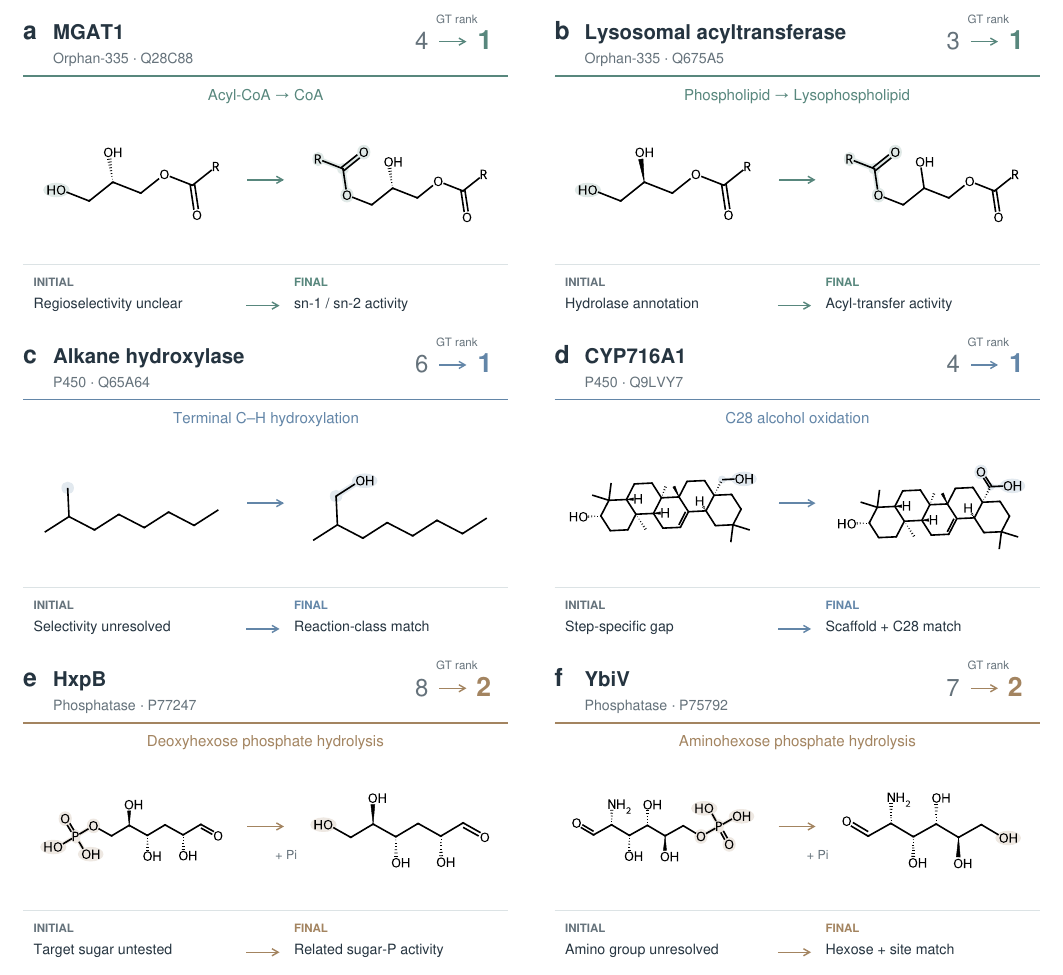}
\caption{\textbf{Evidence-guided reassessment improves the priority of reference-positive candidates across three enzyme retrieval datasets.} \textbf{a,b,} MGAT1 (Q28C88; rank 4 to rank 1) and a lysosomal acyltransferase (Q675A5; rank 3 to rank 1) from Orphan-335. Subsequent assessment strengthened biochemical information concerning monoacylglycerol acceptor-related activity, acyl-CoA dependence, and acyl transfer from phospholipids to alcohol acceptors, respectively. \textbf{c,d,} An alkane hydroxylase (Q65A64; rank 6 to rank 1) and CYP716A1 (Q9LVY7; rank 4 to rank 1) from P450. Reranking emphasized compatibility between reaction class and P450/EC evidence in the former case and consistency among the triterpene scaffold, C28 oxidation site, and carboxyl-containing product in the latter. Previously documented C28 oxidation activity does not, by itself, establish every oxidation step shown in panel d.~\textbf{e,f,} The phosphatases HxpB (P77247; rank 8 to rank 2) and YbiV (P75792; rank 7 to rank 2). Activities involving related sugar-phosphate substrates, metal dependence, and compatibility of phosphate positions supported the ranking changes. Molecular structures were drawn from benchmark SMILES, with highlighted regions indicating the displayed chemical changes. The acyl donors and products in panels a and b are simplified; R denotes an unspecified substituent and does not indicate atom mapping. Pi denotes inorganic phosphate.}
\label{fig:cases}
\end{figure}

Beyond aggregate statistics, we selected six cases in which reference-positive candidates improved in rank across three retrieval settings---Orphan-335, P450, and Phosphatase---to examine how the system reorganizes its assessment of evidence according to different substrates and reaction requirements (Fig.~\ref{fig:cases}a--f). These cases involve acyl transfer, alkane hydroxylation, triterpene C28 oxidation, and sugar-phosphate hydrolysis. They therefore allow us to assess whether RIVER relies only on broad functional labels or can distinguish competing candidates at the level of specific chemical transformations.

For the two acyl-transfer reactions in Orphan-335, the central task of evidence reassessment was to distinguish \textbf{general functional classes} from \textbf{specific acyl acceptors and modes of reaction} (Fig.~\ref{fig:cases}a,b). The initial assessment of MGAT1 had identified its monoacylglycerol acyltransferase-related function, but the sn-1/sn-2 position and substrate scope remained unclear. After further consideration of recorded substrate-activity annotations and acyl-CoA dependence, the reference enzyme moved from rank 4 to rank 1. For Q675A5, the initial annotation was more strongly associated with the hydrolase class; subsequent analysis identified its acyltransferase-related function and prioritized mechanistic compatibility with acyl transfer from phospholipids to alcohol acceptors, moving it from rank 3 to rank 1.

The two P450 cases further demonstrate that \textbf{substrate scaffolds and specific oxidation positions} can provide discriminative evidence among enzymes of the same family (Fig.~\ref{fig:cases}c,d). The initial information for alkane hydroxylase Q65A64 supported general alkane C--H hydroxylation but did not adequately establish the range of accepted branched alkanes. Reassessment strengthened the connection among reaction class, oxidase family, and EC annotation, moving it from rank 6 to rank 1. The analysis of CYP716A1 linked the triterpene scaffold, the C28 position, and the carboxyl-containing product, raising its rank from 4 to 1.

In the phosphatase cases, ranking revisions relied more heavily on \textbf{transferring evidence from related sugar-phosphate substrates and catalytic requirements} (Fig.~\ref{fig:cases}e,f). HxpB was recorded as having sugar- or polyol-phosphatase-related activity; when other sugar-phosphate reactions and divalent metal requirements were considered, it moved from rank 8 to rank 2. Based on hexose-phosphate-related activity and the compatibility of the phosphate position with the target reaction, YbiV moved from rank 7 to rank 2. Both cases show that, in the absence of direct functional annotations for an exact substrate, existing reaction information for related substrates can narrow the candidate range, while unvalidated aspects concerning the target deoxy sugar or amino sugar must remain explicit.

These cases complement the aggregate ablations and evidence profiles: ablation analyses measure the effect of each module on overall metrics, whereas the six specific reactions show which substrates, sites, and functional conflicts are documented in the reasoning traces. RIVER's interpretability thus takes the form of \textbf{inspectable evidence chains with clearly defined information boundaries}.

\subsection{Summary of results}\label{summary-of-results}

Starting from retrospective retrieval of experimentally validated reactions, RIVER prioritized all four target enzymes among 667 \emph{E. coli} candidate proteins and achieved higher MRR than the compared general-purpose coding-agent combinations and existing domain methods. Further results on public bidirectional retrieval benchmarks and enzyme-family retrieval indicate that this advantage is not restricted to a single class of enzymatic reaction. Module ablations show that reaction reasoning and candidate summarization play different important roles across tasks; evidence statistics and representative cases demonstrate that the system can examine compatibility among substrates, reaction sites, functional annotations, and pathway context while recording missing evidence and potential conflicts. Compared with outputs containing candidate rankings alone, this traceable organization of biochemical evidence makes RIVER better suited to proposing testable enzyme--reaction hypotheses and connecting computational prioritization with subsequent experimental work.

\section{Methods}\label{methods}

\subsection{RIVER: an evidence-guided framework for bidirectional enzyme--reaction retrieval}\label{river-an-evidence-guided-framework-for-bidirectional-enzymereaction-retrieval}

RIVER (Reaction--enzyme Inference via Evidence-guided Retrieval) is a training-free multi-agent framework for prioritizing enzyme--reaction associations. It combines biochemical reasoning, retrieved annotations, and computational analyses in a two-stage workflow comprising evidence-guided prescreening and reflective reranking (Fig.~\ref{fig:framework}). Here, \emph{training-free} means that RIVER does not train or fine-tune model parameters on the evaluated retrieval datasets.

The functions of several RIVER agents are as follows. The InitRank Agent establishes the initial candidate order. The Reconsider Agent revisits supporting rationales and identifies evidence gaps or alternative biochemical explanations, while the Contradiction Agent checks for potentially inconsistent evidence. BioChem Expert provides computational evidence, including family, domain, and catalytic-site information where available, to support reassessment. The EvidenceCheck Agent reviews the consistency of the collected evidence and remaining gaps, and the FinalRank Agent generates the final ranking. Together, these components allow candidates to be reassessed as new support or conflicts emerge.

We formulate the two retrieval directions using a query \(q\) and the corresponding candidate pool \(C(q)\):

\[
C(q)=\begin{cases}
E, & q=r\quad\text{(reaction-to-enzyme retrieval)},\\
R, & q=e\quad\text{(enzyme-to-reaction retrieval)}.
\end{cases}\label{eq:1}\tag{1}
\]

Here, \(E\) denotes the specified set of candidate enzymes, and \(R\) denotes the specified set of candidate reactions. In reaction-to-enzyme retrieval, a reaction is evaluated against candidate proteins; in enzyme-to-reaction retrieval, a protein is evaluated against candidate reactions. Both directions assess the biochemical compatibility of enzyme--reaction pairs, differing in which member of the pair is held fixed.

RIVER generates an ordered candidate list accompanied by candidate-specific evidence and unresolved questions. The order represents relative priority within the supplied candidate pool, rather than a calibrated probability of catalytic activity. Multiple candidates may be compatible with a query; candidates lacking positive benchmark labels are not assumed to be experimentally inactive.

\subsection{Evidence-guided prescreening}\label{evidence-guided-prescreening}

Prescreening assembles reaction-level and protein-level information and reduces the candidate pool before detailed reassessment. For each enzyme--reaction pair \((e,r)\), the framework constructs structured textual records:

\[
z_r=F_{\mathrm{reaction}}(r),\qquad z_e=F_{\mathrm{annotation}}(e).\label{eq:2}\tag{2}
\]

Here, \(z_r\) describes the queried transformation, while \(z_e\) summarizes available protein evidence. These records are textual descriptions rather than learned numerical embeddings.

\textbf{ReactionAnalysis Agent.} The ReactionAnalysis Agent interprets the supplied reaction description, including the identities and chemical structures of substrates and products where available. It summarizes the transformation type, substrate features, relevant bond changes, and potential cofactor or cosubstrate requirements. Mechanistic interpretations are treated as hypotheses that guide candidate assessment; requirements that cannot be determined from the available information are retained as unresolved questions.

\textbf{EnzymeAnnotate Agent.} The EnzymeAnnotate Agent organizes available protein information, including UniProtKB annotations, enzyme-family assignments, domain descriptions, and previously documented catalytic activities. Retrieved reaction records, including Rhea entries where available, provide additional chemical context. Candidate summaries preserve distinctions among documented activities, broader family-level compatibility, and inferred substrate scope. Missing annotations are treated as evidence gaps rather than evidence that an enzyme cannot catalyze the reaction.

\textbf{Filter Agent.} The Filter Agent jointly evaluates reaction descriptions and candidate summaries to retain candidates for subsequent ranking:

\[
C_1(q)=F_{\mathrm{filter}}\!\left(q,C(q),\{z_r,z_e\}\right),\qquad C_1(q)\subseteq C(q).\label{eq:3}\tag{3}
\]

Filtering considers biochemical compatibility, including reaction classes, substrate features, and catalytic requirements. Its aim is to reduce the number of candidates requiring detailed review while retaining plausible associations. Retention is a computational prioritization decision and does not establish catalytic activity; likewise, exclusion is not an experimentally validated negative assignment.

\subsection{Evidence records and initial ranking}\label{evidence-records-and-initial-ranking}

For each retained candidate \(c\in C_1(q)\), RIVER assembles a query-conditioned evidence dossier \(D_{q,c}\). The dossier organizes available reaction descriptions, protein annotations, retrieved records, computational outputs, and agent assessments. It distinguishes supporting observations, biochemical interpretations, potential conflicts, and requests for further validation. Evidence availability may differ across candidates and experimental settings.

The InitRank Agent compares retained candidates using evidence available before reflective review and generates an initial order:

\[
\pi_q^{(0)}=F_{\mathrm{init}}\!\left(q,C_1(q),\{D_{q,c}^{(0)}\}_{c\in C_1(q)}\right).\label{eq:4}\tag{4}
\]

Here, \(\pi_q^{(0)}\) is a ranked list, and \(D_{q,c}^{(0)}\) denotes the evidence available at initial ranking. Candidates are assessed jointly in the context of the query, rather than receiving independent, calibrated probabilities of activity.

\subsection{Reflective review and computational evidence}\label{reflective-review-and-computational-evidence}

Reflective review examines whether the initial ordering is supported by reaction-specific evidence and whether alternative interpretations or unresolved conflicts warrant reassessment. Review may change how existing evidence is interpreted; a revised rationale does not necessarily imply that new external evidence has been retrieved.

\textbf{Reconsider Agent.} The Reconsider Agent reexamines the initial candidate ordering and supporting rationales. It considers missing support, alternative biochemical interpretations, and uncertainties in substrate or product specificity. In particular, it focuses on distinguishing broad reaction-class compatibility from evidence for the specific transformation under consideration. For example, documented turnover of the query substrate is distinguished from preference for another substrate, and formation of an intermediate is distinguished from subsequent conversion.

\textbf{Contradiction Agent.} The Contradiction Agent checks evidence that may challenge a candidate assignment, including incompatible substrate scope, alternative reaction chemistry, cofactor requirements, and inconsistencies among retrieved descriptions. Explicit conflicting observations are distinguished from missing information. In particular, the absence of a documented activity or catalytic feature is not automatically interpreted as a contradiction.

\textbf{BioChem Expert.} When tools and suitable inputs are available, the BioChem Expert module supplements language-based assessments with computational evidence. Such evidence may include sequence-family and domain assignments from HMMER/Pfam~\cite{ref40,ref41} and mechanism or catalytic-site information from M-CSA~\cite{ref43}. These results provide additional constraints on enzyme--reaction compatibility, but family membership or an isolated computational match does not establish activity toward the queried substrate.

Tool-derived support depends on the analyses actually executed and the outputs returned in each experiment. Missing, unavailable, or failed tool outputs do not provide positive biochemical support. Protein structures displayed solely for post hoc visualization are not treated as evidence used during ranking.

\textbf{EvidenceCheck Agent.} The EvidenceCheck Agent reviews the consistency, specificity, and remaining uncertainties of the assembled records. It checks whether rationales are supported by documented evidence, whether apparent conflicts remain unresolved, and whether claims about the exact queried activity extend beyond the available annotations or computational results. This review evaluates the internal support for candidate prioritization; it does not independently validate catalytic activity.

\subsection{Final ranking and traceable outputs}\label{final-ranking-and-traceable-outputs}

The FinalRank Agent compares candidates using evidence available after review and produces the final ordering:

\[
\pi_q^{(\mathrm{final})}=F_{\mathrm{final}}\!\left(q,C_1(q),\{D_{q,c}^{(\mathrm{reviewed})}\}_{c\in C_1(q)}\right).\label{eq:5}\tag{5}
\]

The agent integrates reaction-specific support, computational observations, and unresolved conflicts through evidence-conditioned reasoning, without imposing a fixed weighted sum across evidence categories.

For each reported candidate, the output includes its rank and corresponding evidence summary:

\[
O_q=\left[\left(c,\operatorname{rank}_{\pi_q^{(\mathrm{final})}}(c),D_{q,c}^{(\mathrm{reviewed})}\right)\right]_{c\in\pi_q^{(\mathrm{final})}}.\label{eq:6}\tag{6}
\]

Candidate summaries describe supporting rationales, evidence sources, and potential conflicts or evidence gaps. The recorded traces make the basis for prioritization inspectable while preserving distinctions among retrieved observations, computational predictions, and language-model interpretations.

The same evidence-guided formulation applies to both retrieval directions: reaction-to-enzyme retrieval prioritizes candidate catalysts for a specified transformation, while enzyme-to-reaction retrieval prioritizes possible reactions for a specified protein. The resulting rankings support experimental candidate selection and review of functional annotations, but do not, by themselves, establish catalytic activity or physiological relevance.

\section{Discussion}\label{discussion}

RIVER addresses a practical problem in biochemical discovery: determining which enzyme--reaction hypotheses merit experimental attention when annotations are incomplete and evidence is distributed across different resources. Its central contribution is not merely the use of multiple agents, but the organization of heterogeneous evidence acquisition, biochemical interpretation, review, and candidate ranking into a bidirectional workflow.

The next step is prospective biochemical validation. RIVER can help prioritize candidates for orphan reactions, targeted promiscuity assays, analysis of missing steps in metabolic reconstruction, and enzyme engineering projects. Follow-up experiments should measure substrate specificity, cofactor acceptance, product identity, catalytic rates, and appropriate negative controls; a high rank should not be interpreted as evidence of physiological relevance. Complete query-level outputs, frozen database snapshots, and transparent evaluation inventories would further improve reproducibility~\cite{ref38,ref50}. By presenting candidate rankings alongside traceable evidence and unresolved questions, RIVER provides a framework for connecting computational enzyme retrieval to hypothesis-driven experimental research.

\printbibliography[title={References},heading=bibintoc]
\end{document}